\documentclass{article}
\usepackage{spconf,amsmath,graphicx,algorithm,algpseudocode}

\title{Building Corpora for Single-Channel Speech Separation Across Multiple Domains}
\name{Matthew Maciejewski, Gregory Sell, Leibny Paola Garcia-Perera, Shinji Watanabe, Sanjeev Khudanpur}
\address{Center for Language and Speech Processing \\ Human Language Technology Center of Excellence \\
               The Johns Hopkins University, Baltimore, MD 21218, USA}
\begin{document}
\ninept
\maketitle
\begin{abstract}
To date, the bulk of research on single-channel speech separation has been conducted using clean, near-field, read speech, which is not representative of many modern applications. In this work, we develop a procedure for constructing high-quality synthetic overlap datasets, necessary for most deep learning-based separation frameworks. We produced datasets that are more representative of realistic applications using the CHiME-5 and Mixer 6 corpora and evaluate standard methods on this data to demonstrate the shortcomings of current source-separation performance. We also demonstrate the value of a wide variety of data in training robust models that generalize well to multiple conditions.
\end{abstract}
\section{Introduction}
\label{sec:intro}

{\let\thefootnote\relax\footnotetext{This work has been submitted to the IEEE for possible publication. Copyright may be transferred without notice, after which this version may no longer be accessible.}}

A common situation that arises in audio featuring multiple speakers is that two speakers will inevitably speak at the same time. This can lead to a breakdown of performance in speech technologies such as automatic speech recognition (ASR) and speaker identification, as the models are unable to tease apart the speech from the different sources. Speech separation seeks to solve this problem by producing multiple clean, non-overlapping waveforms from a recording in which multiple speakers are talking at the same time.

The bulk of recent work conducted on speech separation \cite{dc,dpcl2,pit,upit,danet,danet2,RSHN} has been done using the Wall Street Journal (WSJ0) \cite{wsj0} corpus, which is not representative of many speech separation tasks. This dataset consists of read speech of utterances from Wall Street Journal news, recorded in a clean environment on close-talking microphones, which is then synthetically combined to form overlapped speech. In the advent of home assistant devices, a common application situation is one in which conversational speech has been recorded in the home from a far-field microphone. Our goal is to establish high-quality datasets that expand evaluation conditions for speech separation and to demonstrate the shortcomings of current standard speech separation techniques on them.

Since evaluation metrics and model training typically require ground truth, supervised methods are typically limited to using synthetic overlap data, where non-overlapping speech utterances are summed together. As such, creating new speech separation datasets involves selecting a source corpus and combining speech from that corpus in a way that creates high-quality overlapping speech.

For our datasets we used the CHiME-5 \cite{chime} and Mixer 6 \cite{mx6} corpora, which each include multiple microphones, allowing near/far-field comparisons. We created a system to isolate high-quality single-speaker speech regions and developed a principled method for pairing utterances. Finally, we used the utterance-level Permutation Invariant Training (uPIT) method \cite{upit} to investigate speech separation performance on these conditions.

\section{Corpus Selection and Data Preparation}
\label{sec:data}

When working with synthetically overlapped speech, careful considerations must be made in data selection. Not only does the selection of the source corpora play a role, but care also must be taken in how the speech is overlapped so that the resulting dataset meets the needs of the application and does so with efficient use of the source speech.

To ensure this high standard of quality would be met, we devised a process for dataset creation with three main components: corpus selection, data cleanup for single-speaker segmentation, and algorithmic generation of mixture lists. We also are releasing the code and resulting mixture lists used in our experiments\footnote{https://github.com/mmaciej2/kaldi/tree/chime5-single-speaker-generation/egs/chime5/single\_speaker\_generation}\textsuperscript{,}\footnote{https://github.com/mmaciej2/kaldi/tree/mixer6-single-speaker-generation/egs/mixer6/single\_speaker\_generation}.

\subsection{Corpus Selection}
\label{ssec:corpora}

For our experiments, we used the WSJ0, CHiME-5, and Mixer 6 speech corpora.

The WSJ0 corpus was selected due to the pre-existence of a synthetic overlap dataset \cite{dc}, a standard of speech separation evaluation. We did not generate our own WSJ0 dataset, instead using the existing dataset. This dataset has been effectively used in a number of speech separation research experiments, and so its composition was the model for our dataset generation pipeline. For our experiments, the WSJ0 overlap dataset also represents the clean, near-field, read speech condition.

The CHiME-5 corpus was chosen to serve as the most challenging, ``realistic'' condition. The corpus consists of dinner parties recorded with microphone arrays placed around the apartment as well as binaural microphones. This condition resulted in a number of unique challenges in the audio, such as naturally occurring non-speech noises, multiple simultaneous speakers, and time-varying locations.  In support of the CHiME-5 challenge, the dataset was fully transcribed, which proved to be essential for the creation of a dataset for speech separation involving realistic condition speech.  In addition, we were able to generate parallel near-field and far-field datasets with identical utterances from this corpus.

The Mixer 6 corpus was chosen to serve as a middle ground between the WSJ0 and CHiME-5 corpora. Consisting of interviews and read speech recorded with 14 microphones in a constructed recording room, the Mixer 6 corpus allows a similar near- and far-field comparison, but in a more controlled environment with stationary speakers, consistent channel, and relatively minimal noise.

Mixer 6 sessions consist of three components: an interview, read speech, and a phone call. For our mixtures we used only speech from the interview session, as we preferred the spontaneity of the format, while the existence of separate close-microphone recordings for both the interviewer and interviewee simplified the process of extracting single-speaker segments, as described below.

\subsection{Cleanup Methods}
\label{ssec:cleanup}

The quality of a machine learning system is highly dependent on the quality of its training data, a characteristic that is especially true for deep learning. As such, it is critical that the original segments used for constructing synthetic overlap contain speech from only one person---no more, no less.

To accomplish this, we used a pipeline implemented with the Kaldi Speech Recognition Toolkit \cite{kaldi} to clean the single-speaker segmentation. This pipeline consists of three parts: initial segmentation, speech activity detection, and segment verification.

\subsubsection{Initial Single-Speaker Segmentation}
\label{sssec:initseg}

The initial segmentation is the only part of the corpus cleanup to differ between the CHiME-5 and Mixer 6 corpora. The purpose of the initial segmentation is to use the basic corpus properties and annotations to generate regions where we have some base confidence level that there is speech present and that only one person is talking.

For the CHiME-5 corpus, we have full speech transcriptions, so the initial segmentation was done by identifying the regions where only one speaker was marked to have been speaking.

For the Mixer 6 corpus, the initial segmentation was done through simple energy analysis between the lapel microphones of the interviewer and the participant. The energy level in the participant's microphone had to exceed a tuned threshold to be considered speech, and the ratio of energy between the microphones had to exceed a tuned threshold to be considered speech by the participant and only the participant.

\subsubsection{Speech Activity Detection}
\label{sssec:sad}

The second stage in the corpus cleanup pipeline is to run a speech activity detection (SAD) system designed for ASR on top of the initial segmentation, which serves multiple purposes.

The first purpose is to remove non-speech segments, as human annotators can make mistakes or be imprecise about annotations, and the energy-based system can be fooled by things such as rustling of clothes. The second purpose is to break the segmentation up into reasonable ASR-style utterances using natural pauses to split longer regions which could prove undesirable for synthetic overlap.

The SAD system is a Time-Delay Neural Network-based system with statistics pooling \cite{ghahremani2016raw} for long-context. The model was trained on data from the LibriSpeech corpus \cite{librispeech}, perturbed with room impulse responses and additive noise/music from the MUSAN corpus \cite{musan}.

For optimal performance, for a given utterance, the SAD system was run on the corresponding speaker's close-talking microphone, which was then mapped back across all microphones using time-alignment drift annotations in the case of CHiME-5 and relying on the synchronous recording property of Mixer 6.

\subsubsection{Segment Verification}
\label{sssec:segver}

The last stage in the corpus cleanup pipeline is to ensure the final segments are sufficiently high-quality. Part of this includes removing utterances deemed ``too short for use'' (shorter than 1.3 seconds, the minimum length of the WSJ0-based mixture corpus), but this stage primarily is done using speaker verification techniques.

We used x-vectors \cite{xvec}, a deep-neural-network-based embedding for speaker identification, along with a Probabilistic Linear Discriminant Analysis (PLDA) scoring backend \cite{plda1,plda2}, state-of-the-art in the speaker identification task. The models were trained using the VoxCeleb corpus \cite{voxceleb} augmented with music and noise \cite{musan}.

Since speaker embeddings contain information about the voice quality of the enrolled speaker, the system can be used to reject segments with unexpected acoustic properties for that speaker, which might arise from undesirable issues like incorrect speaker labels, previously-missed overlapping speech, or non-speech vocalizations. The procedure was to extract and enroll x-vectors over all speech for a given speaker, extract x-vectors for each individual segment, score the segment x-vectors against their speaker's enrolled x-vector using PLDA, and finally reject segments whose scores were below a given threshold. The threshold was tuned based on qualitative assessment of randomly-selected segments.

\subsection{Mixture List Generation}
\label{ssec:mixgen}

\begin{table}[tb]
\caption{Dataset Statistics}
\begin{minipage}[b]{1.0\linewidth}
  \centering
  \centerline{\includegraphics[width=3.4in]{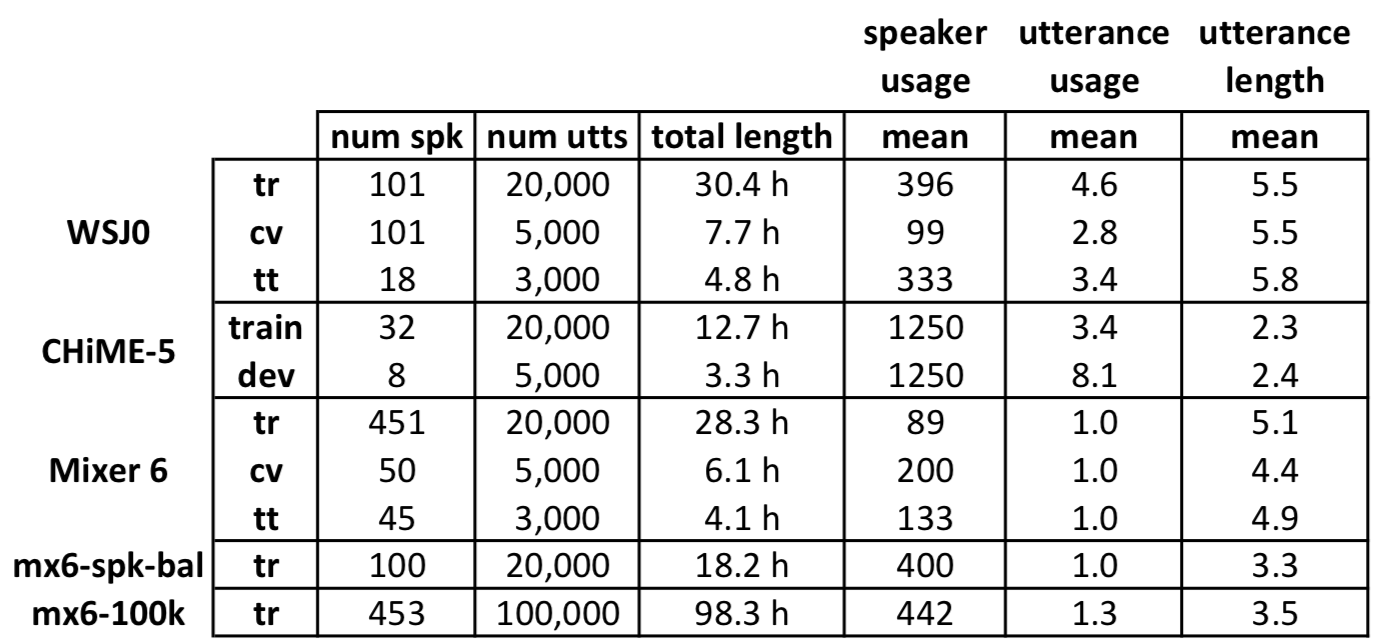}}
\end{minipage}

\label{fig:res1}
\end{table}
\parskip 0truept

For consistency, we generated the mixture datasets by creating lists compatible with the MERL scripts for generating overlap\footnote{http://www.merl.com/demos/deep-clustering/create-speaker-mixtures.zip} that also roughly matched the properties of the WSJ0 mixtures. Despite this, there was a lot of freedom in how to pair utterances from the base corpus to generate mixtures. As a result, we created the mixture lists algorithmically according to a set of desirable criteria selected to maximize the quality of the resulting mixtures given a set of utterances from a corpus.

Mixture lists were constructed based on four criteria. First, we wanted to avoid mixtures of two utterances by the same speaker. Then, we wanted maximal coverage of utterances---for example not using an utterance for a third time if another utterance had only be used once (equivalent to minimizing the maximum amount of data repetition). Next, we wanted to maximize speaker diversity within pairs, i.e. not combining an utterance with two utterances by the same speaker (similar to the previous constraint, except over speaker pair repetition instead of utterance repetition). The final constraint is to pair utterances with utterances of similar length. Since the longer utterance in a pair must be truncated (or the shorter padded), a mismatch in length leads to `wasted' audio, and we thus want to pair long utterances with long utterances and short utterances with short utterances.

\begin{algorithm}[tb]
\begin{minipage}[b]{1.0\linewidth}
\caption{Mixture list generation}\label{algorithm}
\begin{algorithmic}
  \While{$\text{num\_mixes} < \text{target\_mixes}$}
    \State $ \mathcal{S}_1 = \{ utt : \text{usage\_count} (utt) = \min \text{usage\_count} ( \cdot ) \} $
    \State $ u_1 = \arg \max_{u \in \mathcal{S}_1} \text{length}(u) $
    \State $ i \gets 0 $
    \While{$u_1$ not yet matched}
      \State $ \mathcal{S}_2 = \{ utt : \text{usage\_count} (utt) = \min \text{usage\_count} ( \cdot ) + i \} $
      \State $ \mathcal{S}_3 = \{ utt : \text{spk}(utt) \notin  \{ spk : u_1 \text{ previously paired} \} \} $
      \If{$ \mathcal{S}_2 \cap \mathcal{S}_3 \ne \emptyset $}
        \State $ u_2 = \arg \min_{u_2 \in \mathcal{S}_2 \cap \mathcal{S}_3} \mid \text{len}(u_1) - \text{len}(u_2) \mid $
        \State pair $u_1$ and $u_2$, update data structures
      \Else
        \If{$ \mathcal{S}_2 = \emptyset $}
          \State $ \{ spk : u_1 \text{ previously paired} \} \gets \emptyset $
          \State $ i \gets 0 $
        \Else
          \State $ i \gets i+1 $
        \EndIf
      \EndIf
    \EndWhile
  \EndWhile
\end{algorithmic}
\end{minipage}
\end{algorithm}

Since it is computationally infeasible to compute an optimal solution, we used a greedy algorithm, elaborated in Algorithm 1.  The basic structure is to proceed through a priority queue of utterances, constructing sets of eligible utterances to be paired with it and selecting the eligible utterance that is closest in length. Every time an eligible set is empty, we relax the least important constraint. The process is repeated until a target number of mixtures has been generated.

\subsection{Description of Resulting Datasets}
\label{ssec:dsets}

\subsubsection{Single-Speaker Data}
\label{sssec:ssdsets}

The CHiME-5 dataset cleanup resulted in roughly 8 hours of speech across 40 speakers. Speakers on average contributed 327 utterances with a mean length of 2.23 seconds (standard deviation of 0.94), for an average of 12 minutes of speech per speaker.

The Mixer 6 dataset cleanup produced 163 hours of speech from 548 speakers. The average number of utterances per speaker was comparable to CHiME-5 at 341 utterances, though with roughly 50\% more speech on average, with a mean utterance length of 3.13 seconds and standard deviation of 1.44, with an average of 18 minutes of speech per speaker.

The above statistics refer to unique speech, and do not account for the fact that both corpora were recorded on a number of microphones. CHiME-5 was recorded with six 4-microphone arrays in addition to binaural microphones on each of the four participants. Mixer 6 was recorded on 14 microphones of various type and positioning. All of our processing allows for mapping utterances' time marks between all microphones.

We selected two microphone conditions from each corpus for use in our experiments. For the far-field CHiME-5 condition, we selected the first channel of the first microphone array, which is consistent within recordings but not across sessions. For the near-field CHiME-5 condition, we used the binaural microphone for the speaker corresponding to each utterance. For the far-field Mixer 6 condition, we chose channel 9, which is the microphone placed farthest from the speaker. For the near-field condition, we chose channel 2, which is the lapel microphone for the subject.

\subsubsection{Overlapped Speech Data}
\label{sssec:mixdsets}

In constructing the CHiME-5 and Mixer 6 mixture data, an attempt was made to match the WSJ0 mixture data as closely as possible.

The most natural correspondence was to construct train, cross-validation, and test sets of equivalent size (20,000, 5,000, and 4,000 utterances respectively). We chose mixture SNR levels following the same distribution as well. However, because the size of each source corpus varied, the usage of each speaker and utterance varied as well. Comparison of usage statistics are in Table 1. For all conditions, there is no overlap of speakers between data subsets.

For the CHiME-5 data, to increase compatibility with the CHiME-5 ASR evaluation, the data split corresponds with the ASR split. At time of writing, the evaluation transcripts had not yet been released, so cleanup and mixture generation for a test set could not be done. As a result, we evaluated using the cross-validation set generated from the development data and omitted cross-validation.

For the Mixer 6 data, we constructed subsets of 45 and 50 high-quality speakers, roughly gender balanced, to generate testing and cross-validation sets, leaving 451 for the training set. 

In both the CHiME-5 and Mixer 6 mixture datasets, we constructed both near-field and far-field conditions. When doing so, we used identical utterance pairs and mixture weights, as opposed to generating new mixture sets, to reduce the number of confounding factors when comparing performance between near-field and far-field conditions.

In addition, due to the extensive size of the Mixer 6 corpus, the usage statistics of the resulting mixture datasets generated by our algorithm are dissimilar to the WSJ0 mixture dataset. As a result we constructed two additional sets of Mixer 6 mixtures.  In the first, training data was restricted to 100 speakers in order to be `speaker-balanced' to the WSJ0 training data (mx6\_*\_spk-bal). In the second, the total size of the training dataset was increased five-fold to 100,000, resulting in more similar speaker usage.  This allowed us to do an analysis of the trade-off between number of speakers and amount of speech per speaker.

Finally, we constructed a training set of equivalent size by combining of each of the five base datasets (WSJ0, CHiME-5 near and far, Mixer 6 near and far). Two iterations were created. In the first, the combinations were sub-sampled to maintain the size of 20,000 training examples.  In the second, they were fully combined, resulting in 100,000 examples. These sets allowed us to analyze the potential for producing a robust system based on training on a wide variety of properly manicured data.

\section{Speech Separation Experiments}
\label{sec:methods}

\subsection{Separation Technique}
\label{ssec:model}

For evaluation of the data, we used a deep learning mask-based technique called utterance-level Permutation Invariant Training (uPIT) \cite{upit}. The method consists of using a recurrent neural network to produce spectral masks from an input mixture magnitude spectrum, then applying the masks to the mixture spectrogram and inverting to reconstruct the estimated source waveforms.

For computational reasons, we used down-sampled 8 kHz audio. The spectrograms were generated using a short-time Fourier transform (STFT) with a window length of 512 and step of 128. The magnitude spectrum was used for input to the neural network.

For the purposes of our experiments, it was assumed that the input speech was a mixture of two speakers, and so the network was specifically designed to output two masks.  This is not a limitation of the method or uPIT training strategy, but rather a choice for simplifying the experiments.

The network used consists of two 600-node bidirectional LSTM layers followed by a linear layer, with sigmoid output. It is trained with mean squared error loss between the estimated (masked) and ground truth source magnitude spectra:
$$
\mathrm{Loss}(\mathbf{\hat{M}}, \pi) = \frac{1}{B} \sum_{s=1}^S \| \mathbf{\hat{M}}_{\pi(s)} \circ \mathbf{A}_{mix} - \mathbf{A}_s \|_F^2
$$
where $\mathbf{\hat{M}}$ are the estimated masks, $\mathbf{A}$ are the STFT magnitudes, $B$ is the total number of STFT coefficients across sources, and $\hat{\pi}$ is the permutation of output masks that produces the lowest loss:
$$
\hat{\pi} = \arg \min_{\pi \in \mathcal{P}} \sum_{s=1}^S \mathrm{Loss}(\mathbf{\hat{M}}, \pi)
$$

We used the Adam optimizer with a learning rate of 0.001. The models were trained for 200 training epochs, with the final model being chosen based on a minimum loss on the cross-validation set.

\subsection{Evaluation Metric}
\label{ssec:metric}

For evaluation, we used a standard evaluation metric for source separation, signal to distortion ratio (SDR) \cite{sdr}. It measures the overall estimation error of the target sources.

\section{Results}
\label{sec:results}

\begin{table}[tb]
\caption{SDR improvement across varying train and test conditions}
\begin{minipage}[b]{1.0\linewidth}
  \centering
  \centerline{\includegraphics[width=9cm]{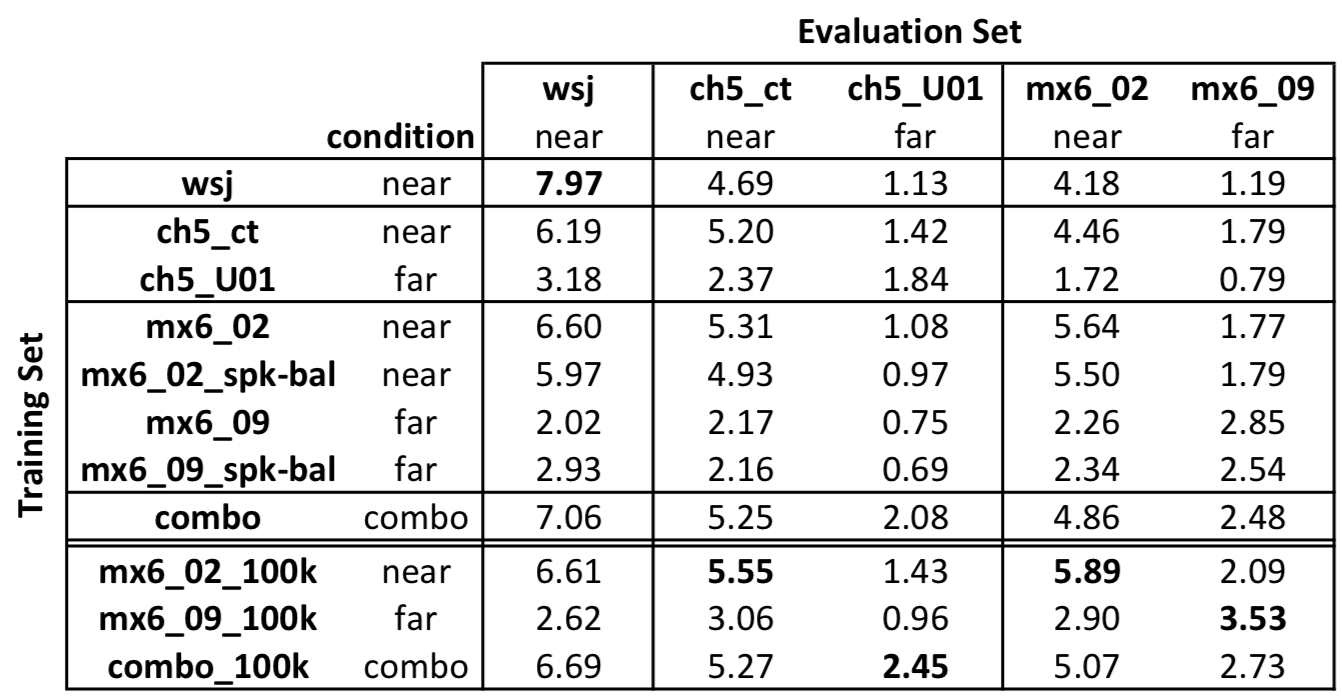}}
\end{minipage}
\end{table}

The results of our experiments are shown in Table 2. The two biggest conclusions are that models do not appear to generalize well and that performance degrades in more realistic far-field conditions.

We see that the best overall separation performance appears in the baseline WSJ0 test condition with the model trained on WSJ0 data. However, there is a clear degradation in performance moving from WSJ0 to more realistic conversational, near-field conditions, and even further degradation in far-field.

We also see that matched conditions between train and test sets typically achieve the greatest performance, but the models do not seem to generalize well to other sets. One exception is that the near-field Mixer 6 model outperformed the near-field CHiME-5 model on the near-field CHiME-5 test condition. It is possible that the higher-quality audio of Mixer 6 produced a better model, though the performance between the two models is close enough that strong conclusions should not be made.  In general, near-field trained systems perform reasonably well on other near-field conditions while degrading in far-field, and vice versa.

Something else worth noting is that the Mixer 6 training set where the number of speakers was balanced to match the WSJ0 dataset performed slightly worse that regular Mixer 6 training set. The small degradation in performance might be due to the overall decrease in total speech present in the dataset, though performance is again close enough that strong conclusions should not be made. However, the absence of noticeable improvements suggests that having depth in speech per speaker does not have much impact over greater diversity in speakers.

The expanded 100,000 Mixer 6 mixture training sets showed some improvement, but the gains were modest and tended to be restricted to matched conditions where the model was already performing well (especially in the far-field case), doing little to improve performance otherwise.

Finally, an interesting result is that the models trained on a combination of training conditions achieved near-optimal performance. Though they performed on average slightly worse than the best system in each test condition, they are the only systems to have success across conditions. Given that the other models performed quite poorly outside of their own domain, this is an important result that demonstrates that diversity of training data is important for producing robust speech separation models.  Furthermore, as with the 100,000 Mixer 6 data, the expanded training set appears to be more valuable in far-field conditions than near-field.

\section{Conclusions}
\label{sec:conclusions}

In this work we have presented and shared a framework for producing new, high-quality synthetic overlap datasets for training and evaluation of source separation technologies, and we have used it to produce additional datasets that expand the domain of single-channel speech separation evaluation from clean, near-field conditions to far-field, realistic conversational speech.

We have demonstrated the value of this work through an analysis of standard speech separation technologies that reveals a large degradation in performance in challenging conditions. We have also shown that including multiple conditions in training data, made possible through this work, greatly improves the robustness of the resulting model, a shortcoming in the technology otherwise.

\section{Acknowledgements}
\label{sec:acknowledgements}

We would like to thank Xuankai Chang for providing comments useful to producing the final version of this manuscript.

\bibliographystyle{IEEEbib}
\bibliography{strings,refs}

\end{document}